\documentclass[journal]{IEEEtran}
\usepackage{graphicx}
\usepackage{amsmath}
\usepackage{array}
\usepackage{booktabs}
\usepackage{tabularx}
\usepackage{multirow}
\usepackage{hyperref}
\usepackage{cite}
\usepackage{placeins}
\usepackage{orcidlink}
\usepackage{threeparttable}  
\usepackage{makecell} 

\title{Prompt-Based Segmentation at Multiple Resolutions and Lighting Conditions using Segment Anything Model 2
}

\author{
    Osher~Rafaeli\raisebox{0.5ex}{\orcidlink{0000-0002-7097-7568}}, 
    Tal~Svoray\raisebox{0.5ex}{\orcidlink{0000-0003-2243-8532}}, 
    Roni~Blushtein-Livnon\raisebox{0.5ex}{\orcidlink{0000-0002-3493-4894}}
    and~Ariel~Nahlieli\raisebox{0.5ex}{\orcidlink{0009-0001-0633-1842}}
    \thanks{O. Rafaeli, R. Blushtein-Livnon and A. Nahlieli are with the Department of Environmental, Geoinformatics and Urban Planning Sciences, Ben-Gurion University of the Negev, Israel (e-mail: osherr@post.bgu.ac.il; livnon@bgu.ac.il; arielnah@post.bgu.ac.il).}
    \thanks{T. Svoray is with the Department of Environmental, Geoinformatics and Urban Planning Sciences, and the Department of Psychology, Ben-Gurion University of the Negev, Israel (e-mail: tsvoray@bgu.ac.il).}%
}

\IEEEaftertitletext{\vspace{-2\baselineskip}}

\begin{document}

\maketitle
\begin{abstract}
This paper provides insights on the effectiveness of the zero shot, prompt-based Segment Anything Model (SAM) and its updated versions, SAM 2 and SAM 2.1, along with the non-promptable conventional neural network (CNN), for segmenting solar panels in RGB aerial remote sensing imagery. The study evaluates these models across diverse lighting conditions, spatial resolutions, and prompt strategies. SAM 2 showed slight improvements over SAM, while SAM 2.1 demonstrated notable improvements, particularly in sub-optimal lighting and low resolution conditions. SAM models, when prompted by user-defined boxes, outperformed CNN in all scenarios; in particular, user-box prompts were found crucial for achieving reasonable performance in low resolution data. Additionally, under high resolution, YOLOv9 automatic prompting outperformed user-points prompting by providing reliable prompts to SAM. Under low resolution, SAM 2.1 prompted by user points showed similar performance to SAM 2.1 prompted by YOLOv9, highlighting its zero shot improvements with a single click. In high resolution with optimal lighting imagery, Eff-UNet outperformed SAMs prompted by YOLOv9, while under sub-optimal lighting conditions, Eff-UNet, and SAM 2.1 prompted by YOLOv9, had similar performance. However, SAM is more resource-intensive, and despite improved inference time of SAM 2.1, Eff-UNet is more suitable for automatic segmentation in high resolution data. This research details strengths and limitations of each model and outlines the robustness of user-prompted image segmentation models.
\end{abstract}


\section{Introduction}

Remote sensing involves gathering information about the terrestrial Earth’s surface, water bodies, and the atmosphere via airborne or satellite platforms, serving a wide range of applications \cite{sabins2020remote}. Recent advances in computer vision (CV) facilitated automatic pixel-level classification analyses and greatly improved image segmentation \cite{MA2019166}.

Convolutional Neural Networks (CNNs) were found successful for image segmentation as they typically obtain semantic representations using stacked convolutions and pooling operations restoring image size and pixels location by upsampling \cite{DBLP:journals/corr/RonnebergerFB15}. Based on U-shaped networks, various  encoder–decoder designs were applied \cite{zhu2017deep}. For example, the novel architecture Eff-UNet combines the effectiveness of compound scaled pre-trained EfficientNet, as the encoder for feature extraction, with the UNet decoder for reconstructing fine-grained segmentation maps \cite{9150621}. 

CNN-based image segmentation depends on training quality in terms of diversity, representativeness, and model's ability to address similar tasks after training \cite{SHARMA2020566}. Recently, Prompt-based Vision Transformer (ViT) models, which enable users to control outputs interactively during prediction, became  commonly-used \cite{Luddecke_2022_CVPR}. These models allow correcting errors along the process and visualizing ongoing predictions \cite{MAZUROWSKI2023102918}.

Prompt-based models can enhance prediction from zero shot (where a model classifies classes it hasn't encountered during training), allowing generalization based on learning underlying data concepts/relationships and interpreting unseen data \cite{pratt2023doesplatypuslooklike}.  Meta FAIR gained progress with zero shot prompt-based image segmentation during 2023-24. The segment anything model (SAM) demonstrates this approach by conducting image segmentation with minimal human intervention. It only requires a bounding box or a clicked point for a prompt \cite{kirillov2023segment}. Furthermore, when SAM is guided by YOLO-generated box prompts \cite{7780460}, the entire segmentation pipeline can become end-to-end automatically \cite{zhu2024activelearningenabledlowcost}. Since SAM was released in 2023, the scientific community has expressed large interest in implementing it for RS purposes, integrating SAM in various applications \cite{osco2023segmentmodelsamremote}. Particularly, throughout 2024, engineers invested resources to develop adapters for improving SAM accuracy to handle remotely sensed data \cite{wang2023samrsscalingupremotesensing}. 

SAM was used to segment various objects on the Earth surface, such as vegetation, buildings, geological features, and water bodies. This was done directly from satellite, airborne, and drone imagery, without additional fine-tuning \cite{shankar2023semantic, osco2023segmentmodelsamremote}. However, in a zero shot setting, SAM is prone to over-segmentation, particularly when object boundaries are amorphous and unclear, often requiring multiple prompts to achieve accurate results \cite{osco2023segmentmodelsamremote}. While fine-tuning SAM for specific objects improves segmentation accuracy \cite{10493013}, the zero shot capabilities of previous SAM versions remain widely adopted due to its robustness and interactive user interface, supporting diverse tasks. SAM is already integrated into GIS and is commonly employed in RS workflows \cite{Wu2023}.

SAM 2 was introduced in late July 2024, and its updated version SAM 2.1, in late September 2024, extending promptable visual segmentation. According to Meta research, SAM 2 and SAM 2.1 outperform the previous version in terms of accuracy and inference time. SAMs are trained and tested on ground-level photographs and videos \cite{ravi2024sam}, thus, they encounter challenges when applied to RS imagery due to their low resolution, high background-target imbalance, and inconsistency in lighting and atmospheric conditions \cite{wang2023samrsscalingupremotesensing}. Thus, although SAM's latest versions introduce improvements, their benefits have yet to be applied and quantified in RS.

Our \textit {aim} here, is to compare three segmentation approaches: (1) The conventional fully automatic CNNs, including trained conventional Eff-UNet with pre-trained on ImageNet EfficientNet encoder and classic U-Net; (2) CV-promptable SAM, SAM 2, and SAM 2.1, prompted with bounding boxes generated by YOLOv9 object detection model; and (3) semi-automatic approach implementing user-promptable SAMs prompted by user created bounding boxes and one-click centroid points. Within this comparison, we outline three objectives: (1) to quantify the zero-shot capabilities of SAMs in RS data; (2) to investigate the effects of different prompt strategies on model performance in SAMs, including user-created points, user-created bounding boxes, and computer vision-generated bounding boxes using YOLO; and (3) to evaluate model performance across three airborne datasets with low resolution (0.25 m) and high resolution (0.15 m) under both optimal and suboptimal lighting conditions.

\section{Materials and Methodology}

\subsection{Study area and dataset}

\begin{figure}[htbp!]
    \centering
    \includegraphics[width=1\linewidth]{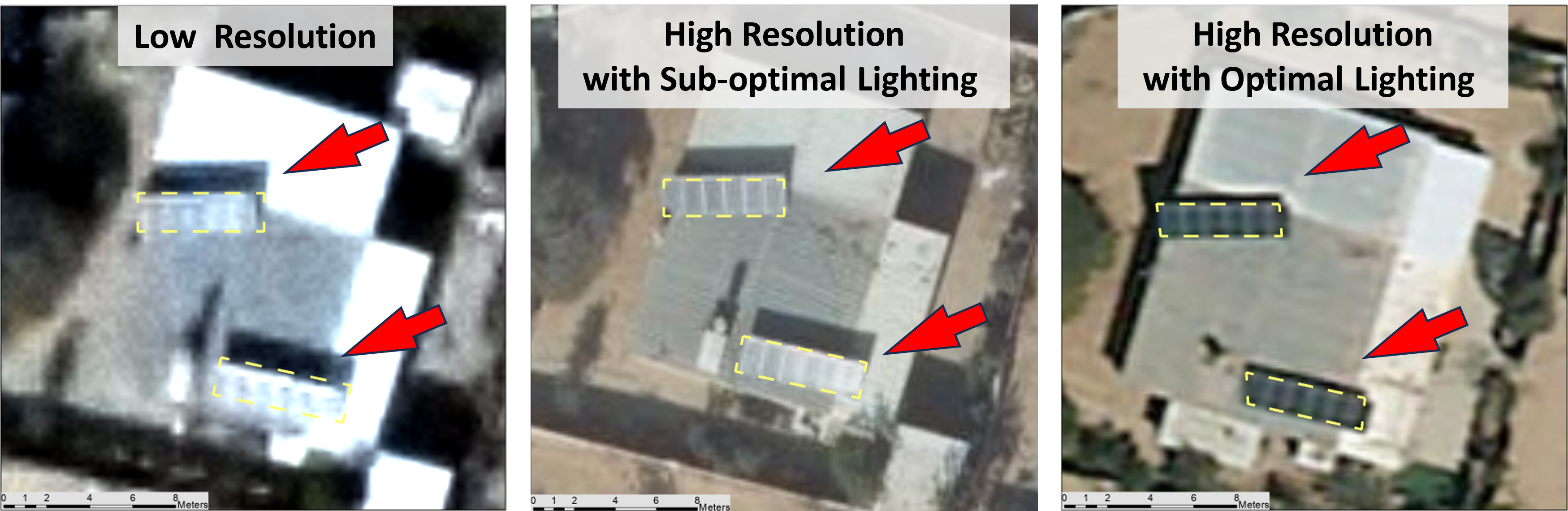}
    \caption{PV cells captured in images with different resolutions and lighting conditions: High resolution (0.15 m) and low resolution image (0.25 m), highlighting the increased difficulty in segmentation as resolution decreases and lighting condition becomes sub-optimal.} \label{resolutions}
\end{figure}

We selected a heterogeneous rural area, near the city of Be'er Sheva, Israel, to pose a challenge to models' performance in segmenting sparse targets such as small-scale photovoltaic (PV) cells. The area is covered with scattered settlements, agricultural fields, and  rangelands. Airborne images were obtained spanning 3 years with varying resolutions and lighting conditions: 2022 with high resolution (0.15 m) and optimal lighting conditions; 2017 with high resolution (0.15 m) and sub-optimal light conditions; and 2015 for low resolution (0.25 m). These images were used to assess models' performance across quality. Overall $ \approx {3000} $ PV cells were identified and labeled in each airborne dataset (Fig. \ref{resolutions}).

Following the airborne survey, aerial photo interpretation and annotation were conducted to generate precise masks of PV cells. These masks serve as the basis for: (1) training the CNN models, Eff-UNet, U-Net, and YOLO; (2) generating user prompts in a semi-automatic approach, which includes bounding boxes and centroid points; and (3) estimating model performance on unseen data.

\subsection{Eff-UNet}
Convolutional neural network progressively reduces input image resolution to obtain the high-level feature map representing the original image, and the decoder module consists of a set of layers that upsamples the feature map of the encoder to recover spatial information. We implemented the encoder path with cutting-edge set of CNN EfficientNet, introduced in 2019 by Google AI research \cite{DBLP:journals/corr/abs-1905-11946}. For decoder, U-Net allows the network to propagate context information to higher resolution layers. We used EfficientNetB1 pre-trained on ImageNet utilizing transfer learning to adapt it for PV cells segmentation task. Additionally, we implemented a simple U-Net to generate baselines to the models.

\subsection{YOLO}
You Only Look Once (YOLO) proposes end-to-end neural network object localization and detection models. The speed and accuracy of YOLO models made them widespread in scientific and industrial applications \cite{wang2024yolov9}. Here, YOLO models provided bounding-box generation for PV cells in aerial images. These automatic bounding boxes were then used to prompt both SAM and SAM 2. During the intermediate tests, we trained and evaluated the three latest state-of-the-art YOLO models: YOLOv8, YOLOv9 and YOLOv10 \cite{wang2024yolov10, wang2024yolov9} to identify the most suitable model for PV cells segmentation. YOLOv9 was chosen as it outperformed YOLOv8 and YOLOv10 in detecting the small PV cells in challenging scenarios. YOLOv9 achieved an IoU score of 0.630 and an F1-score of 0.724, while v8 and v10 achieved an IoU of 0.602 and 0.611 and an F1-score of 0.692 and 0.703, respectively. 

\subsection{Training and implementation details}

Images of PV cells and corresponding ground truth masks, extracted from a reference "gold standard dataset," created through a controlled annotation process conducted by human RS experts, were processed into patches of 256 x 256 pixels, and randomly divided into training, validation, and test sets, with 70\% (\textasciitilde 2100 PV cells) of the labels used to train the Eff-UNet, YOLOv9 and U-Net models, 10\% (\textasciitilde 300 PV cells) for validation, and 20\% (\textasciitilde 600 PV cells) for testing. The models were trained with 75 epochs, a batch size of 4 images, and an initial learning rate of 0.001. The learning rate decayed exponentially. Dice Loss Function was employed for UNet and Eff-UNet. Other configurations and parameter settings for YOLO were set to default.
Our implementation utilized high-performance computational resources to ensure efficient training and inference of the segmentation models. Specifically, we employed the Ultralytics v8.3.22 framework, which includes the YOLO studied versions \cite{Jocher_Ultralytics_YOLO_2023} with PyTorch-2.3.1 for YOLO and SAMs, and Keras module from TensorFlow library for U-Net and EfficientNet. All was running on Python 3.9.15 with an NVIDIA RTX A4000 GPU.

\subsection{Segment anything models}
SAM models were used in their pre-trained zero shot large (ViT-L) configurations with default parameters for inference (without fine-tuning or using adapters). The user and YOLO prompts were used as input to SAMs. All prompting points are used as input together as foreground prompts to SAM models, while for bounding boxes, masks were extracted for each bounding box separately, and then, all image predictions were combined. If more than a single object occurred in a single image, we combined prediction by choosing pixels with higher foreground probability (thresholded to 0.5).

\subsection{Model performance metrics}

To assess models' performance, we employed the widely recognized evaluation metrics in Equation 1: 

\setlength{\abovedisplayskip}{1pt}
\setlength{\belowdisplayskip}{1pt}

\begin{equation}
\begin{aligned}
\text{F1 Score} &= \frac{TP}{TP + \frac{1}{2}(FP + FN)} \\
\text{IoU} &= \frac{|A \cap B|}{|A \cup B|}
\end{aligned}
\label{eq:metrics}
\end{equation}

where TP, TN, FP, and FN denote, respectively, the true positive, true negative, false positive, and false negative values, and \(A\) and \(B\) represent the predicted and ground truth segments, respectively. IoU denotes Intersection over Union.

\section{Results}
\begin{figure}[htbp!]
    \centering
    \includegraphics[width=1\linewidth]{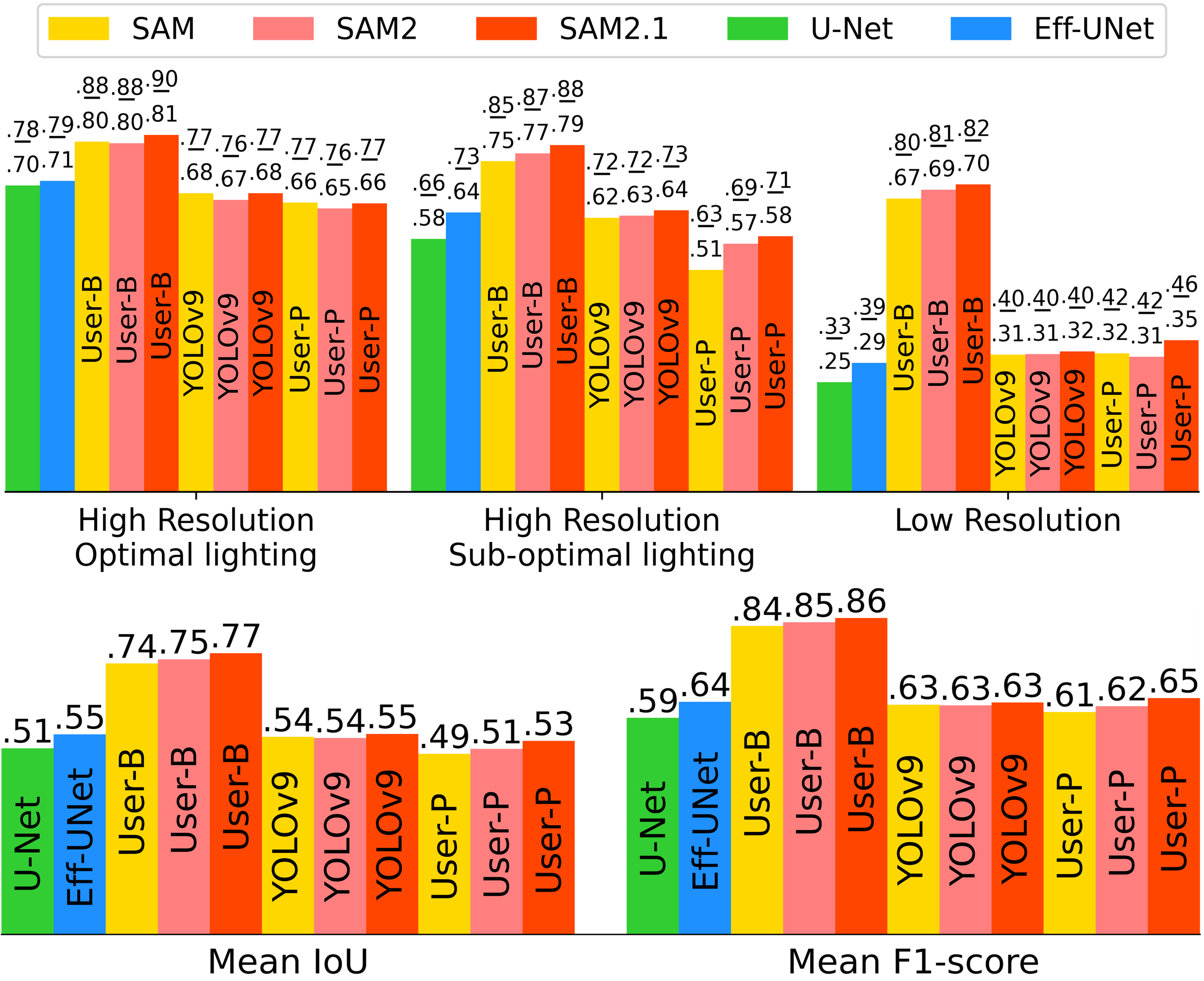}
    \caption{Upper chart: IoU and F1-score (\underline{underlined}) results across all three studied datasets. Lower chart: mean IoU and F1-score across the datasets.}   \label{iou_f1_combined}
\end{figure}

Performance of segmentation models of high resolution images was evaluated under optimal and sub-optimal lighting conditions while segmentation of low resolution images was evaluated under optimal light conditions only (Table \ref{tab:comparison} and Fig. \ref{iou_f1_combined}). Models include: (1) SAMs with user points and user boxes, and YOLOv9 generated Boxes; and (2) CNN non-promotable models, Eff-UNet, and U-Net baseline models.

\begin{table*}[htbp]
    \centering
    \caption{Comparison of performance metrics across the three datasets and their mean performance}
    \label{tab:comparison}
    \scriptsize
    \begin{tabular}{>{\centering\arraybackslash}p{3.5cm} >{\centering\arraybackslash}p{1cm} | >{\centering\arraybackslash}c >{\centering\arraybackslash}c >{\centering\arraybackslash}c | >{\centering\arraybackslash}c >{\centering\arraybackslash}c >{\centering\arraybackslash}c | >{\centering\arraybackslash}c >{\centering\arraybackslash}c >{\centering\arraybackslash}c | >{\centering\arraybackslash}c | >{\centering\arraybackslash}c}
    \toprule
    \textbf{} & \textbf{} & \multicolumn{3}{c|}{\textbf{User Points}} & \multicolumn{3}{c|}{\textbf{User Box}} & \multicolumn{3}{c|}{\textbf{Box (YOLOv9)}} & \multirow{2}{*}{\textbf{U-Net}} & \multirow{2}{*}{\textbf{Eff-UNet}} \\ 
    \cmidrule{3-11}
    & & \textbf{SAM} & \textbf{SAM2} & \textbf{SAM2.1} & \textbf{SAM} & \textbf{SAM2} & \textbf{SAM2.1} & \textbf{SAM} & \textbf{SAM2} & \textbf{SAM2.1} & & \\ 
    \Xhline{2\arrayrulewidth}
    \multirow{2}{*}{\textbf{High Res. with Opt. Lighting}} & IoU & \textbf{0.660} & 0.646 & 0.658 & 0.799 & 0.795 & \textbf{0.814} & \textbf{0.681} & 0.666 & \textbf{0.681} & 0.699 & \textbf{0.709} \\ 
    & F1  & \textbf{0.770} & 0.761 & \textbf{0.770} & 0.882 & 0.882 & \textbf{0.895} & \textbf{0.767} & 0.758 & \textbf{0.767} & 0.780 & \textbf{0.790} \\ 
    \Xhline{2\arrayrulewidth}
    \multirow{2}{*}{\textbf{High Res. with Sub-Opt. Lighting}} & IoU & 0.506 & 0.566 & \textbf{0.583} & 0.754 & 0.772 & \textbf{0.791} & 0.625 & 0.630 & \textbf{0.642} & 0.577 & 0.637 \\ 
    & F1  & 0.629 & 0.693 & \textbf{0.707} & 0.853 & 0.867 & \textbf{0.880} & 0.720 & 0.724 & \textbf{0.732} & 0.664 & \textbf{0.732} \\ 
    \Xhline{2\arrayrulewidth}
    \multirow{2}{*}{\textbf{Low Res.}} & IoU & 0.316 & 0.308 & \textbf{0.346} & 0.669 & 0.689 & \textbf{0.701} & 0.313 & 0.314 & \textbf{0.320} & 0.250 & 0.294 \\ 
    & F1  & 0.425 & 0.419 & \textbf{0.462} & 0.795 & 0.810 & \textbf{0.820} & 0.396 & 0.397 & \textbf{0.402} & 0.331 & 0.386 \\ 
    \Xhline{2\arrayrulewidth}
    \multirow{2}{*}{\textbf{Mean}} & IoU & 0.494 & 0.507 & \textbf{0.529} & 0.741 & 0.752 & \textbf{0.769} & 0.540 & 0.537 & \textbf{0.548} & 0.509 & \textbf{0.547} \\ 
    & F1  & 0.608 & 0.624 & \textbf{0.646} & 0.843 & 0.853 & \textbf{0.865} & 0.628 & 0.626 & \textbf{0.634} & 0.592 & \textbf{0.636} \\ 
    \Xhline{2\arrayrulewidth}
    \end{tabular}
\end{table*}

\subsection{High resolution with optimal lighting conditions}
SAM 2.1 outperformed both SAM and SAM 2 when prompted by user bounding boxes, achieving an IoU of 0.814 and F1-score of 0.895. A decline in performance was observed in point-prompted SAMs that achieved an IoU of 0.660 and F1-score of 0.770, with similar overall results for the three SAM versions. Among the fully automatic models, Eff-UNet outperformed in high resolution with optimal lighting with an IoU of 0.709 and F1-Score of 0.790.

\subsection{High resolution with sub-optimal lighting conditions}
Under sub-optimal lighting conditions, we identified significant variability in performance across models. SAM's prompted by the user box performed consistently better than other configurations. Among SAMs, SAM 2.1 prompted by user boxes notably outperformed with an IoU of 0.791 and F1-Score of 0.880, while SAM 2 achieved an IoU of 0.772 and F1-Score of 0.867 and SAM achieved an IoU of 0.754 and F1-Score of 0.853. However, when prompted by points, SAM 2.1 and SAM 2 markedly surpass SAM, with an IoU of 0.583 and an F1-Score of 0.707 and an IoU of 0.566 and an F1-Score of 0.693 correspondingly, while SAM dropped significantly, achieving an IoU of 0.506 and F1-Score of 0.629. SAM prompted by points has resulted in partially segmented PV cells and falsely part of the background. SAM 2.1 demonstrated this bias less and, therefore, better performed. Among the automatic models, Eff-UNet achieved an IoU of 0.637 and an F1-Score of 0.732. Similarly, the performance of SAM 2.1 prompted by YOLOv9 boxes aligned closely with IoU of 0.642 and F1-Score of 0.732. 

\subsection{Low resolution}
SAMs with manually generated box prompts demonstrate reasonable accuracy even using low resolution images, while SAM 2.1 outperformed, achieving an IoU=0.701 and F1-score=0.820. By contrast to user box prompts, SAMs with user-generated points struggled to extract PV boundaries correctly. Still, SAM 2.1 outperformed, achieving a low IoU=0.346 and F1-Score=0.462. The performance of all automatic models declined under low resolution conditions. Even Eff-UNet, achieving best results at higher resolutions, struggled to effectively learn patterns and extract these features from unseen tested data. YOLOv9 also struggled to supply reliable boxes as prompts to SAMs.

\subsection{Mean performance across resolutions and light conditions}
When averaging the accuracy measures across resolutions and light conditions (Table \ref{tab:comparison}), SAM 2.1 with user box prompts showed consistently high performance across resolutions, indicating its robustness in various conditions. Achieving an IoU of 0.769 and an F1-Score of 0.865. Similarly, SAM 2.1 outperformed SAM and SAM 2 with points prompt that achieved an IoU of 0.529 and F1-Score of 0.646. SAMs prompted by YOLOv9 outperformed user single-click prompts. Eff-UNet and SAM 2.1 promoted by YOLOv9 showed similar performance: IoU=0.548 and F1-Score=0.634 (Table \ref{tab:comparison}).

\subsection{Computation resources}
The automatic configuration of SAMs consists of three stages: (1) training an object detection model; (2) detecting objects (bounding boxes) with this model; and (3) SAM inference. Stages 1 and 2 can alternatively use user-provided prompts. YOLOv9e requires 2.10 GB of GPU memory and 3 minutes per training epoch, with inference time=97.5 ms for a four-PV-cell image, in addition to YOLOv9e detection inference time, SAM segmentation inference time is 593 ms per detected bounding box for SAM (299 ms for SAM2 and SAM2.1). The inference time scales linearly with the number of bounding boxes. Using user prompts, the total processing time is 593 ms for SAM and 299 ms for SAM 2, prompted by a batch of point prompts or a single bounding box. In comparison, Eff-UNet, required only 1.38 GB of GPU memory, completed each training epoch in 43 seconds and inferred in 47 ms.

\section{Conclusion}
\begin{figure}[htbp!]
    \centering
    \includegraphics[width=1\linewidth]{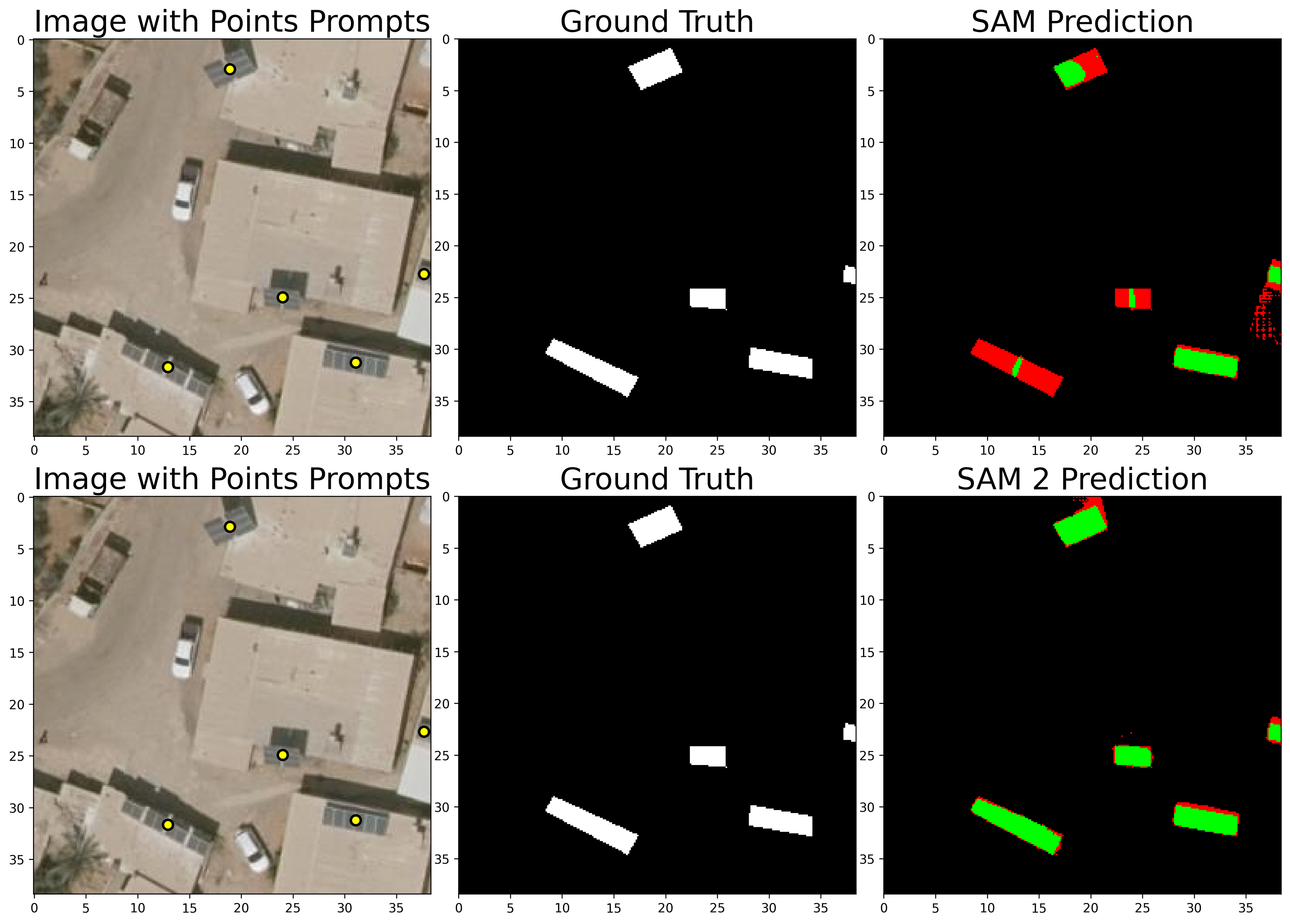}
    \caption{SAM Vs. SAM 2: Under sub-optimal lighting imagery, SAM 2 segments PVCs in full while SAM segments them partially, or falsely (Green: TP, Red: FP \& FN).}
    \label{sam_sam2_1}
\end{figure}

\begin{figure}[htbp!]
    \centering
    \includegraphics[width=1\linewidth]{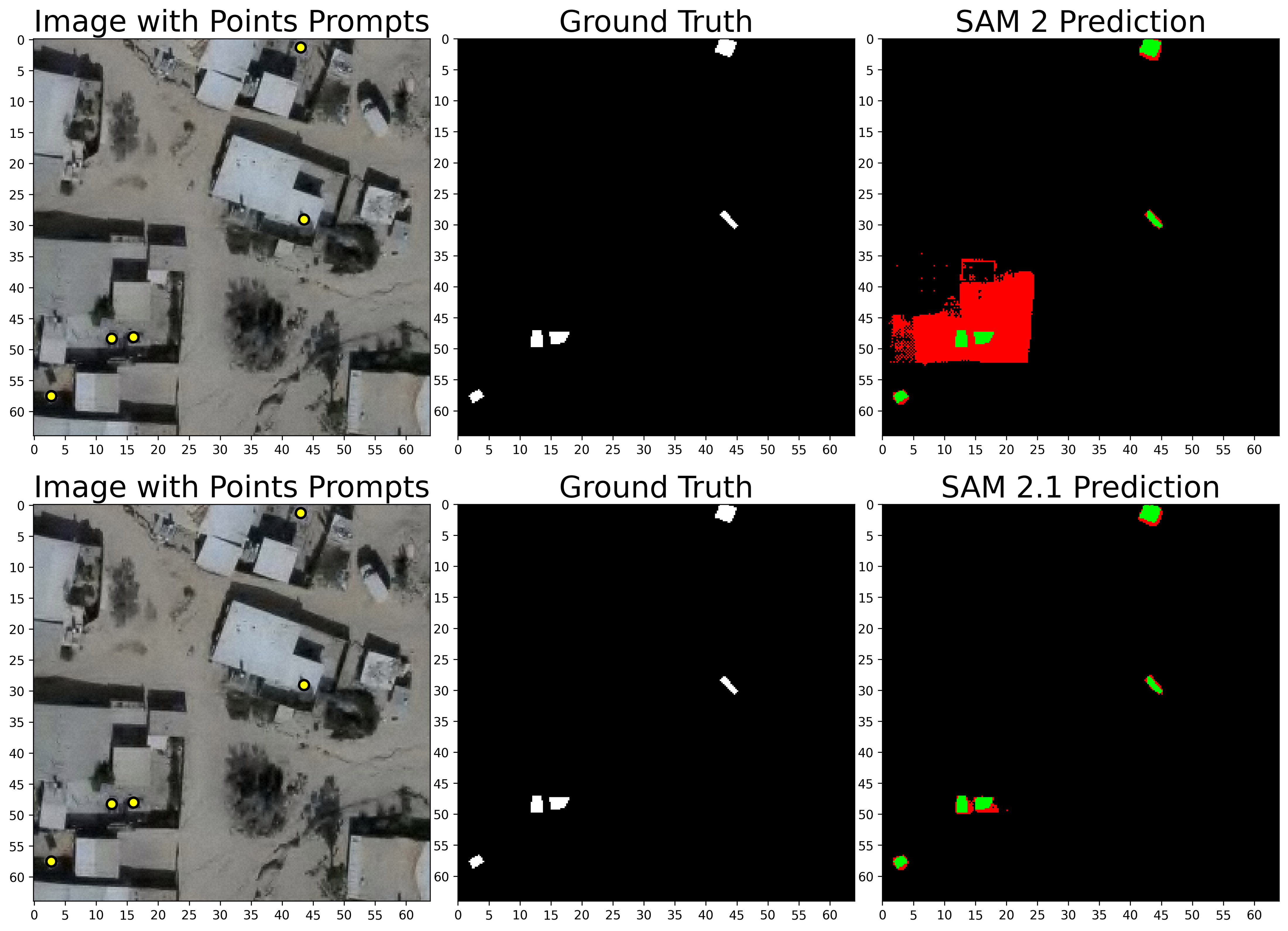}
    \caption{SAM 2 Vs. SAM 2.1 under low resolution: Shadow are clustered with small PVCs segmented by SAM 2 as PVCs (Red: FP \& FN) whereas SAM 2.1 segments correctly.}
    \label{sam2.1vssam2}
\end{figure}

Evaluating segmentation model performance requires examining both accuracy and the resources invested. For example, human annotation usually achieves highest accuracy, yet, it is very labor-intensive. The following insights consider these trade-offs, emphasizing new prompt-based segmentation models that utilize expert prompts to minimize reliance on precise manual annotation:

(1) SAMs zero shot performance: SAM 2 demonstrated improvements over SAM, with better segmentation accuracy, particularly in high resolution imagery under sub-optimal lighting conditions where SAM often partially segmented PV cells or mistakenly included background elements (Fig. \ref{sam_sam2_1}). SAM 2.1 further enhanced these advancements, showing even higher zero shot performance, including for small PV cells in low resolution imagery where SAM 2 often struggled (Fig. \ref{sam2.1vssam2}). These improvements can be attributed to SAM 2's updated architecture, which was expanded further in SAM 2.1 and trained on the larger SA-V dataset, allowing for more precise segmentation in complex scenarios \cite{ravi2024sam}.

\begin{figure}[htbp!]
    \centering
    \includegraphics[width=1\linewidth]{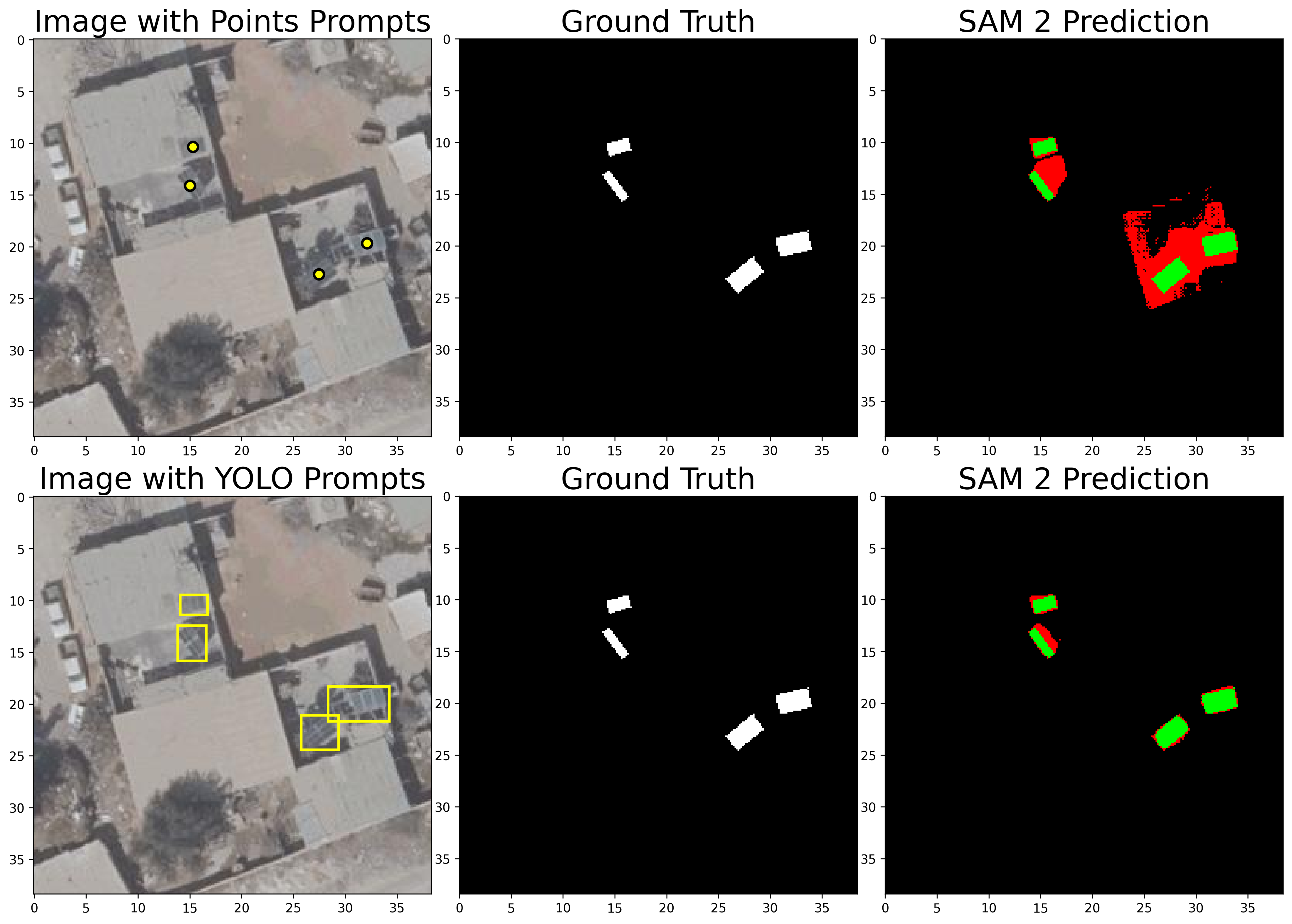}
    \caption{Prompts: under sub-optimal lighting conditions YOLO exceeds single click prompts. boxes lowering FPR by defining the background regions. (Green: TP, Red: FP \& FN).}
    \label{YOLOvsPOINTS}
\end{figure}

(2) Prompt strategies: using user-defined boxes outperformed all SAMs across all three datasets. The performance of YOLO boxes and user points, however, varied depending on dataset. At high resolution, both strategies perform comparably; yet, in suboptimal lighting conditions, YOLO-predicted boxes yielded fewer false positives due to more accurate foreground and background delineation (Fig. \ref{YOLOvsPOINTS}). At low resolution, single-click user surpass YOLO prompts, underscoring strengths of user-based prompts in poor resolution settings and highlighting YOLO's efficiency limitations. 

\begin{figure}[htbp!]
    \centering
    \includegraphics[width=1\linewidth]{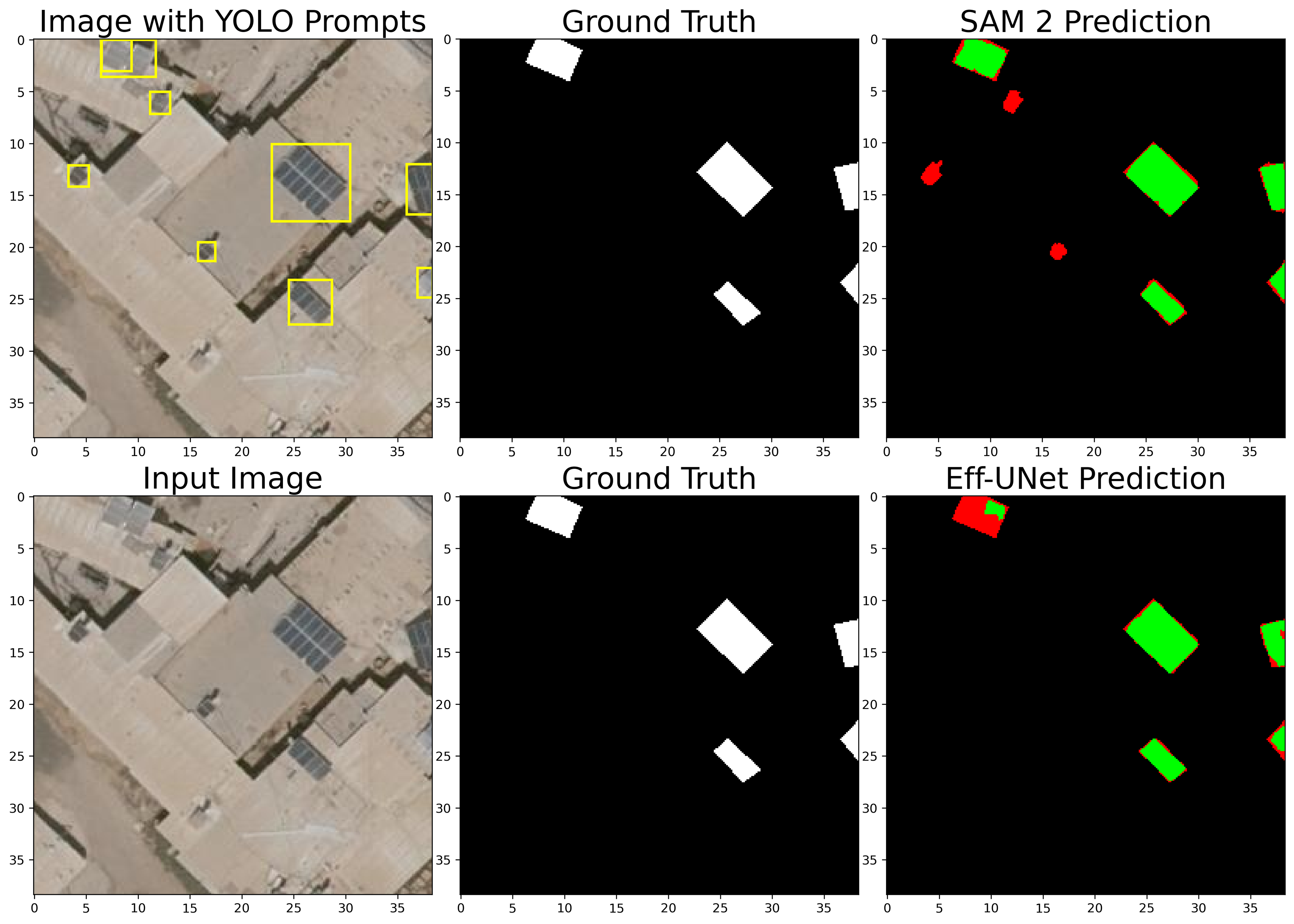}
    \caption{Auto-segmentation: Eff-UNet exceeds SAMs \& YOLO. Incorrect boxes affect many pixels (Green: TP, Red: FP \& FN).}
    \label{yolo_vs_eff}
\end{figure}

(3) Automatic segmentation: automatic segmentation performance highly depended on dataset resolution. As a result, Eff-UNet and SAMs, prompted by YOLO, perform efficiently at high resolution, particularly under optimal lighting conditions. However, their accuracy declined under sub-optimal lighting and lower resolution. At high resolution, Eff-UNet outperforms SAMs prompted by YOLO in accuracy. This result underscored the robust learning capabilities of CNNs for well-defined objects. In contrast, SAM, which relies on user prompts, is heavily influenced by input quality; when using an imperfect CV object detection model, each false positive box may lead to significant segmentation errors across numerous pixels. If no true positive object are detected, SAM does not generate any segmentation for those objects. Eff-UNet, on the other hand, lacks this dependency on prompts, resulting in  more consistent performance (Fig. \ref{yolo_vs_eff}).

Overall, SAM 2.1 introduces substantial improvements that could significantly reduce the annotation workload compared with earlier versions. Despite these improvements, when using high resolution datasets, with well-defined objects, CNN-based automatic segmentation remains preferable over SAMs prompted by YOLO. CNN use allows a single training process on well-defined data and subsequent segmentation, whereas SAMs and YOLO synergy is more resource-intensive, involving multiple stages. Conversely, under varying resolutions and challenging light conditions, where fully automatic models struggle and human-guided prompts become necessary, SAM models offer significant advantages. However, note that single-click prompt in SAM 2.1 remains limited in complex scenarios, and may require multiple user interactions for optimal accuracy. Further research is recommended to validate these observations on less-defined objects and across various satellite and drone datasets.

\FloatBarrier
\section*{Acknowledgment}

We thank the Israel Science Foundation (ISF), grant 299/23; the Ministry of Agriculture Chief Scientist, grant 16-17-0005; "A Partnership for Sustainability" between Ben-Gurion University of the Negev and the Hebrew University of Jerusalem; and the Negev Scholarship by the Kreitman School of Ben-Gurion University for supporting Osher Rafaeli's PhD studies.

\bibliographystyle{ieeetr} 
\bibliography{references}

\end{document}